%% file: main.tex
\documentclass{article}

\usepackage{microtype}
\usepackage{graphicx}
\usepackage{subcaption}
\usepackage{booktabs}
\usepackage{hyperref}

\usepackage[preprint]{conf}

\usepackage{hyperref}
\usepackage{url}
\usepackage{indentfirst}
\usepackage{algorithm}
\usepackage{algorithmic}
\usepackage{amsmath,amsfonts}
\usepackage{algorithmic}
\usepackage{array}
\usepackage{textcomp}
\usepackage{stfloats}
\usepackage{url}
\usepackage{verbatim}
\usepackage{graphicx}
\usepackage{times}
\usepackage{epsfig}
\usepackage{graphicx}
\usepackage{amsmath}
\usepackage{amssymb}
\usepackage{multirow}
\usepackage{booktabs}
\usepackage{amsthm,amsmath,amssymb}
\usepackage{amsmath}
\usepackage{cases}
\usepackage{bm}
\usepackage{color}
\usepackage{mathrsfs}
\usepackage{CJK}
\usepackage[most]{tcolorbox}
\usepackage[table]{xcolor}
\usepackage{mdframed}
\usepackage{tcolorbox}
\usepackage{enumitem}
\usepackage{algorithmic}
\usepackage{amsmath}
\usepackage{graphicx}
\usepackage{balance}

\usepackage[capitalize,noabbrev]{cleveref}

\theoremstyle{plain}

\theoremstyle{definition}

\theoremstyle{remark}

\usepackage[textsize=tiny]{todonotes}

\begin{document}

\twocolumn[
  \icmltitle{Reinforced Curriculum Pre-Alignment for Domain-Adaptive VLMs}
  \icmlsetsymbol{equal}{*}

  \begin{icmlauthorlist}
    \icmlauthor{Yuming Yan}{equal,comp}
    \icmlauthor{Shuo Yang}{equal,sch}
    \icmlauthor{Kai Tang}{comp}
    \icmlauthor{Sihong Chen}{comp}
    \icmlauthor{Yang Zhang}{comp}
    \icmlauthor{Ke Xu}{comp}
    \icmlauthor{Dan Hu}{comp} \\
    \icmlauthor{Qun Yu}{comp}
    \icmlauthor{Pengfei Hu}{comp}
    \icmlauthor{Edith C.H. Ngai}{sch}
  \end{icmlauthorlist}

  \icmlaffiliation{comp}{Tencent}
  \icmlaffiliation{sch}{The Univerisity of Hong Kong}
  \icmlcorrespondingauthor{Shuo Yang}{shuoyang.ee@gmail.com}

  \vskip 0.3in
]

\printAffiliationsAndNotice{\icmlEqualContribution}

\begin{abstract}

Vision-Language Models (VLMs) demonstrate remarkable general-purpose capabilities but often fall short in specialized domains such as medical imaging or geometric problem-solving. Supervised Fine-Tuning (SFT) can enhance performance within a target domain, but it typically causes catastrophic forgetting, limiting its utility as a general AI agent. The central challenge, therefore, is to adapt VLMs to new domains while preserving their general-purpose capabilities.
Continual pretraining is effective for expanding knowledge in Large Language Models (LLMs), but it is less feasible for VLMs due to prohibitive computational costs and the unavailability of pretraining data for most open-source models. This necessitates efficient post-training adaptation methods. Reinforcement learning (RL)–based approaches such as Group Relative Policy Optimization (GRPO) have shown promise in preserving general abilities, yet they often fail in domain adaptation scenarios where the model initially lacks sufficient domain knowledge, leading to optimization collapse.
To bridge this gap, we propose \textbf{R}einforced \textbf{C}urriculum \textbf{P}re-\textbf{A}lignment (\textbf{RCPA}), a novel post-training paradigm that introduces a curriculum-aware progressive modulation mechanism. In the early phase, RCPA applies partial output constraints to safely expose the model to new domain concepts. As the model’s domain familiarity increases, training gradually transitions to full generation optimization, refining responses and aligning them with domain-specific preferences. This staged adaptation balances domain knowledge acquisition with the preservation of general multimodal capabilities.
Extensive experiments across specialized domains (e.g., OpenI for medical imaging and Geo170K for geometry) and general benchmarks (e.g., COCO Captions) validate the effectiveness of RCPA. Results show that RCPA achieves domain-specific performance competitive with SFT while significantly outperforming SFT in retaining general multimodal understanding, establishing a practical pathway toward building high-performing and domain-adaptive VLMs. 

\end{abstract}

\input{TEX/1.Introduction}
\input{TEX/2.Review}
\input{TEX/3.Preliminary}
\input{TEX/4.Method}
\input{TEX/5.Experiment}
\input{TEX/6.Conclusion}

\balance
\bibliography{reference}
\bibliographystyle{conf}

\end{document}

%% file: TEX/1.Introduction.tex
\section{Introduction}

Vision-Language Models (VLMs)~\citep{Qwen2.5-VL, Qwen2-VL, Qwen-VL, wang2025internvl3_5, zhu2025internvl3, chen2024expanding, wang2024mpo,gao2024mini, liu2024llavanext, liu2023improvedllava} have achieved remarkable progress in unifying visual perception with natural language understanding, enabling powerful capabilities across tasks such as image captioning, visual question answering (VQA), and multimodal dialogue~\citep{Qwen2.5-VL, Qwen2-VL, Qwen-VL}. Large-scale pre-training on web-scale multimodal corpora equips these models with broad general-purpose competencies, making them versatile foundation models. However, real-world deployment often requires domain adaptation, where VLMs must acquire specialized knowledge (e.g., medical, scientific, or industrial contexts) while retaining their general reasoning abilities. This dual requirement—learning new domain knowledge without forgetting existing skills—poses a central challenge for post-training adaptation. 

Current post-training adaptation strategies~\citep{Gekhman2024DoesFL,Guo2023FinetuningSF,Hu2021LoRALA,Lambert2025ReinforcementLF,Rafailov2023DirectPO} are generally categorized into two primary paradigms: Supervised Fine-Tuning (SFT) and Reinforcement Learning (RL). SFT facilitates direct knowledge injection by performing imitation learning on target-domain data. While effective for domain-specific expertise transfer, SFT frequently induces catastrophic forgetting~\citep{Shenfeld2025RLsRW}, where specialized signals overwrite established general-purpose capabilities. To mitigate this degradation, various approaches ~\citep{Li2016LearningWF,Kirkpatrick2016OvercomingCF,Rebuffi2016iCaRLIC} have been proposed, ranging from incremental learning ~\citep{Li2016LearningWF,Kirkpatrick2016OvercomingCF,Rebuffi2016iCaRLIC} to Test-Time Adaptation (TTA)~\citep{NEURIPS2023_cdd06402,Niu2023TowardsST,Yang2024AVF}. These methodologies typically employ one of three mechanisms: parameter-efficient fine-tuning, regularization-based knowledge preservation, or exemplar replay from source domains. However, these designs are often ill-suited for large-scale Vision-Language Models (VLMs). Specifically, data replay is frequently rendered infeasible by stringent privacy, storage, or licensing constraints, while regularization-based methods often settle for a suboptimal trade-off, failing to achieve peak performance in both domain specialization and general capability retention simultaneously. Conversely, preference-based RL frameworks, such as Group Relative Policy Optimization (GRPO)~\citep{Shao2024DeepSeekMathPT}, utilize KL-regularized optimization to safeguard model generality. Nevertheless, the efficacy of these methods hinges on the implicit assumption that the pre-trained model already possesses non-trivial domain knowledge~\citep{Wang2025ReinforcementLF}. When this prerequisite is unmet—a common scenario in low-resource or highly specialized domains—the model fails to generate sufficiently high-quality trajectories, inevitably leading to optimization collapse.

The distinct limitations of SFT and RL-based adaptation underscore a fundamental challenge: neither paradigm, in isolation, can concurrently ensure optimization stability, effective knowledge transfer, and the robust preservation of general-purpose capabilities. To address this bottleneck, we propose that domain adaptation should be conceptualized as a progressive rather than a monolithic process. Under this framework, the model must first undergo pre-alignment to establish a stable, foundational grounding in the target domain's concepts. This is subsequently followed by reinforcement-alignment, which leverages richer preference signals to refine model behavior and ensure high-fidelity task performance.

In this paper, we propose \textbf{R}einforced \textbf{C}urriculum \textbf{P}re-\textbf{A}lignment (\textbf{RCPA}), a novel post-training framework that unifies domain knowledge acquisition with preference alignment under a curriculum-driven design. RCPA introduces a progressive modulation mechanism that transitions smoothly from constrained imitation to full generation optimization, avoiding optimization collapse while mitigating forgetting. This framework is built upon GRPON (GRPO for Non-Deep-Thinking Models), which adapts the RL backbone specifically for non-reasoning VQA-style tasks. To further enhance adaptability, RCPA incorporates two curriculum-inspired modules: Curriculum Progress Perception (CPP) and Curriculum Difficulty Perception (CDP). CPP regulates answer-prefix injection and reward threshold scheduling to bootstrap stable signals in early training, while CDP dynamically prioritizes difficult samples to maximize learning benefits and prevent overfitting. Extensive experiments on domain-specific VQA benchmarks demonstrate that RCPA not only achieves superior domain alignment but also preserves broad general-purpose multimodal capabilities, outperforming existing SFT- and RL-based approaches.

% Our main contributions are summarized as follows:
% \begin{itemize}[leftmargin=1em]
%     \item We identify the inherent limitations of existing adaptation paradigms and highlight the necessity of a progressive, curriculum-driven strategy for domain-adaptive VLM alignment.
%     \item  We propose RCPA, a two-phase post-training paradigm that integrates pre-alignment and reinforcement alignment with a progressive modulation mechanism.
%     \item We design CPP and CDP modules that provide curriculum-driven guidance, enabling robust domain knowledge acquisition while safeguarding general competencies.
%     \item We conduct extensive experiment on in-domain and out-of-domain benchmarks, results show that RCPA achieves domain-specific performance competitive with SFT and significantly outperform SFT in retaining the model’s original general multimodal understanding.
% \end{itemize}

% \\\indent (1) We identify the inherent limitations of existing adaptation paradigms and highlight the necessity of a progressive, curriculum-driven strategy for domain-adaptive VLM alignment.
% \\\indent (2) We propose RCPA, a two-phase post-training paradigm that integrates pre-alignment and reinforcement alignment with a progressive modulation mechanism.
% \\\indent (3) We design CPP and CDP modules that provide curriculum-driven guidance, enabling robust domain knowledge acquisition while safeguarding general competencies.
% \\\indent (4) We conduct extensive experiment on in-domain and out-of-domain benchmarks, XX

%% file: TEX/2.Review.tex
\section{Related Works}

\subsection{Vision Language Models (VLMs)}
Vision Language Models (VLMs)~\citep{Qwen2.5-VL, Qwen2-VL,Qwen-VL,wang2025internvl3_5,zhu2025internvl3,chen2024expanding,wang2024mpo,gao2024mini,liu2024llavanext,liu2023improvedllava} have significantly advanced cross-modal intelligence by integrating text and image modalities, progressing through three key phases. In the foundational phase, early models like CLIP~\citep{Radford2021LearningTV} and ViT-BERT~\citep{Li2021TowardsAU} bridged the modal gap between text and images, enabling tasks like zero-shot transfer and visual grounding. Subsequent models such as ALBEF~\citep{Li2021AlignBF} and FLAVA~\citep{Singh2022FLAVAUO} further refined alignment techniques for better semantic consistency. The second phase focused on enhancing general capabilities through instruction tuning, with models like LLaVA~\citep{Liu2023VisualIT} and Flan-V5~\citep{Chung2022ScalingIL} improving cross-modal reasoning and task handling. Recent developments include InternVL~\citep{wang2025internvl3_5,zhu2025internvl3,wang2024mpo}, an open-source Vision Language Models (VLMs) series, and the Qwen series~\citep{Qwen2.5-VL, Qwen2-VL, Qwen-VL}, both pushing the boundaries of multimodal understanding with advanced visual encoders and innovative techniques for handling high-resolution images, multimodal rotation, and tool usage. Despite these advances, VLMs continue to struggle with domain-specific adaptation. In specialized settings such as medical imaging or scientific problem-solving, they often fail to recognize domain-specific concepts or adapt to nuanced visual features. To this end, post-training methods have emerged as a promising direction, offering lightweight yet effective mechanisms for adapting VLMs to specialized domains without retraining from scratch.

\subsection{Post-training for VLMs}

Post-training techniques, primarily SFT and RL, have been central to adapting VLMs to domain-specific tasks~\citep{kumar2025llmposttrainingdeepdive, chu2025sft, lai2025med, li2025drive}. SFT enables task-specific learning but often leads to catastrophic forgetting of the general knowledge learned during pre-training, especially when fine-tuning is performed on domain-specific data~\citep{duan2024cityllava, dong2025scalable,chen2024efficiency}. Parameter-efficient approaches such as QLoRA~\citep{Dettmers2023QLoRAEF}, LoRA~\citep{Hu2021LoRALA}, Adapters~\citep{Hu2023LLMAdaptersAA}, and Prompt/Prefix Tuning~\citep{Lester2021ThePO, Li2021PrefixTuningOC} alleviate computational burdens by updating only a subset of parameters, yet they remain prone to overfitting and limited transferability. In contrast, RL-based methods such as Group Relative Policy Optimization (GRPO)~\citep{Shao2024DeepSeekMathPT}, Domain Adaptive Policy Optimization (DAPO)~\citep{Yu2025DAPOAO}, and Group Sequence Policy Optimization (GSPO)~\citep{Zheng2025GroupSP} enhance adaptability by leveraging dynamic feedback and optimizing sequential decision-making. These methods are effective in preserving general capabilities by regularizing the model’s output with reward signals, but they assume that the pre-trained model already possesses non-trivial domain knowledge. Traditional RL approaches such as Proximal Policy Optimization (PPO)~\citep{Schulman2017ProximalPO}, which require training both the policy and critic models~\citep{Lambert2025ReinforcementLF}, impose high computational costs. Recent alternatives like GRPO and Direct Preference Optimization (DPO)~\citep{Rafailov2023DirectPO} reduce this burden by removing the need for a separate critic, thereby simplifying training. Nevertheless, existing RL paradigms still struggle to adapt efficiently when the model begins with limited domain expertise. The commonly adopted ``SFT-then-RL” pipeline~\citep{Shao2024DeepSeekMathPT} partially alleviates this issue by stabilizing the reward signal in early training. However, recent findings from CHORD~\citep{Zhang2025OnPolicyRM} reveal a fundamental flaw: SFT disrupts the pretrained model’s internal structures, causing temporary degradation of general capabilities, while subsequent RL fails to recover domain adaptation—often performing worse than direct RL. These limitations underscore the pressing need for more efficient and scalable post-training strategies that can jointly achieve domain specialization and general capability preservation.

%% file: TEX/3.Preliminary.tex
\section{Preliminary}
\subsection{Problem Formulation}
Consider a pre-trained VLM denoted as $\pi_{\text{pre}}$, which exhibits strong general-purpose multimodal capabilities. We are given a target domain-specific dataset $\mathcal{D}_{\text{target}} = \{(\mathbf{x}_i, y_i)\}_i^N$, where $\mathbf{x}_i = (\text{image}_i, \text{prompt}_i)$ represents a multimodal input (image paired with a task prompt) and $y_i$ is the ground-truth response containing domain-specific knowledge. The objective is to adapt $\pi_{\text{pre}}$ into a domain-adapted model $\pi_{\theta}$ through a post-training procedure, such that $\pi_{\theta}$ achieves high performance on $\mathcal{D}_{\text{target}}$ while maximally preserving the general capabilities of $\pi_{\text{pre}}$. This defines the fundamental challenge in domain-adaptive VLM alignment: how to integrate novel domain knowledge without forgetting previously acquired knowledge.
\subsection{Limitations of Existing Methods}
\textbf{Supervised Fine-Tuning (SFT).} SFT adapts model parameters $\theta$ by maximizing the likelihood of expert demonstrations in $\mathcal{D}_{\text{target}} = \{(\mathbf{x}_i, y_i)\}_{i=1}^N$. Its objective is:
\begin{equation}
\mathcal{J}_{\text{SFT}}(\theta) = \mathbb{E}{(\mathbf{x}, y) \sim \mathcal{D}_{\text{target}}} \left[ \sum_{t=1}^{|y|} \log \pi_{\theta}(y_t \mid \mathbf{x}, y_{<t}) \right],
\label{eq:sft_objective}
\end{equation}

where $y_{<t}$ denotes the prefix tokens of $y$. While SFT effectively injects domain-specific knowledge via imitation learning, it relies exclusively on supervised labels—though this reliance does not directly cause catastrophic forgetting. Instead, the core driver is the distributional gap between SFT and pretraining data. When target domain data (for SFT) differs substantially from pretraining data, SFT-induced retraining aligns the model with the target domain distribution, leading to misalignment with pretraining data and subsequent catastrophic forgetting. Previously acquired capabilities are overwritten by domain-specific information, undermining VLMs' generalization. 

\textbf{Group Relative Policy Optimization (GRPO).} GRPO and other Preference-based RL methods attempt to align models with human preferences while mitigating forgetting through regularization. The GRPO incorporates a KL-divergence penalty to constrain the updated policy $\pi_{\theta}$ from deviating excessively from a reference policy $\pi_{\theta_{\text{ref}}}$ (typically the initial pre-trained model $\pi_{\theta_{\text{pre}}}$). Formally, given a input $x$ and a group of $G$ responses $O=\{o_1,o_2,\dots,o_G\}$ sampled from the old policy $\pi_{\theta_{\text{old}}}$. The GRPO objective maximizes the expected clipped advantage for each token $o_{i,t}$ in response $o_{i}$:

\begin{equation}
\begin{aligned}
\mathcal{J}_{\text{GRPO}}(\theta) &= \mathbb{E}_{x \sim P(X), \{o_i\}_{i=1}^G \sim \pi_{\theta_{\text{old}}}} \Biggl[ \frac{1}{G} \sum_{i=1}^{G} \frac{1}{|o_i|} \sum_{t=1}^{|o_i|} \\
&\quad \min \Biggl( \rho_{i,t}(\theta) \hat{A}_{i,t}, \text{clip} \left( \rho_{i,t}(\theta), 1 - \varepsilon, 1 + \varepsilon \right) \hat{A}_{i,t} \Biggr) \\
&\quad - \gamma \mathbb{D}_{\text{KL}}\left[ \pi_{\theta} \| \pi_{\text{ref}} \right] \Biggr],
\end{aligned}
\label{eqn:grpo_obj}
\end{equation}

\noindent where the $\rho_{i,t}(\theta) = \frac{\pi_{\theta}(o_{i,t} | x, o_{i,<t})}{\pi_{\theta_{\text{old}}}(o_{i,t} |x, o_{i,<t})}$ is the importance sampling ratio, $\epsilon$ is the clip factor, $\gamma$ controls the strength of KL regularization, and $\hat{A}_{i,t} = \frac{r_i - \text{mean}(\mathbf{r})}{\text{std}(\mathbf{r})}$ is the standardized advantage for token \( o_{i,t} \), computed from the group rewards \( \mathbf{r} = \{r_1, r_2, \dots, r_G\} \). 

GRPO alleviates forgetting by combining preference-based optimization with KL regularization. However, it presupposes that the pretrained model already holds non-trivial knowledge of the target domain. If the initial model $\pi_{\text{pre}}$ possesses limited knowledge of the target domain, it cannot generate responses of sufficient quality to yield informative reward signals, which is also known as \textit{optimization collapse}.

\vspace{-0.3cm}
\subsection{Motivation for RCPA}
\label{ssec:motivation}

Domain-adaptive alignment of VLMs entails a dual objective: injecting novel domain knowledge while preserving general-purpose capabilities. This poses a substantial challenge for existing post-training paradigms. On one hand, SFT is effective at incorporating domain knowledge but inevitably induces catastrophic forgetting of general skills. On the other hand, RL-based methods such as GRPO emphasize preserving broad competencies through regularization, yet often suffer from optimization collapse when the model lacks sufficient prior knowledge of the target domain. Consequently, neither purely supervised nor purely reinforcement-based adaptation can simultaneously ensure stability, effective domain transfer, and robust generalization. This limitation motivates a progressive adaptation strategy, wherein the model is first pre-aligned to safely acquire foundational domain concepts and subsequently reinforcement-aligned to refine its behavior with richer preference signals. Building on this perspective, we propose Reinforced Curriculum Pre-Alignment (RCPA)—a paradigm that introduces a progressive modulation mechanism to dynamically coordinate the training objective, enabling a smooth transition from constrained imitation learning to full reward-driven optimization as the model’s domain familiarity evolves.

%% file: TEX/4.Method.tex
\section{Method}
\subsection{Overview of RCPA}

\begin{figure*}
    \centering
    \includegraphics[width=0.95\linewidth]{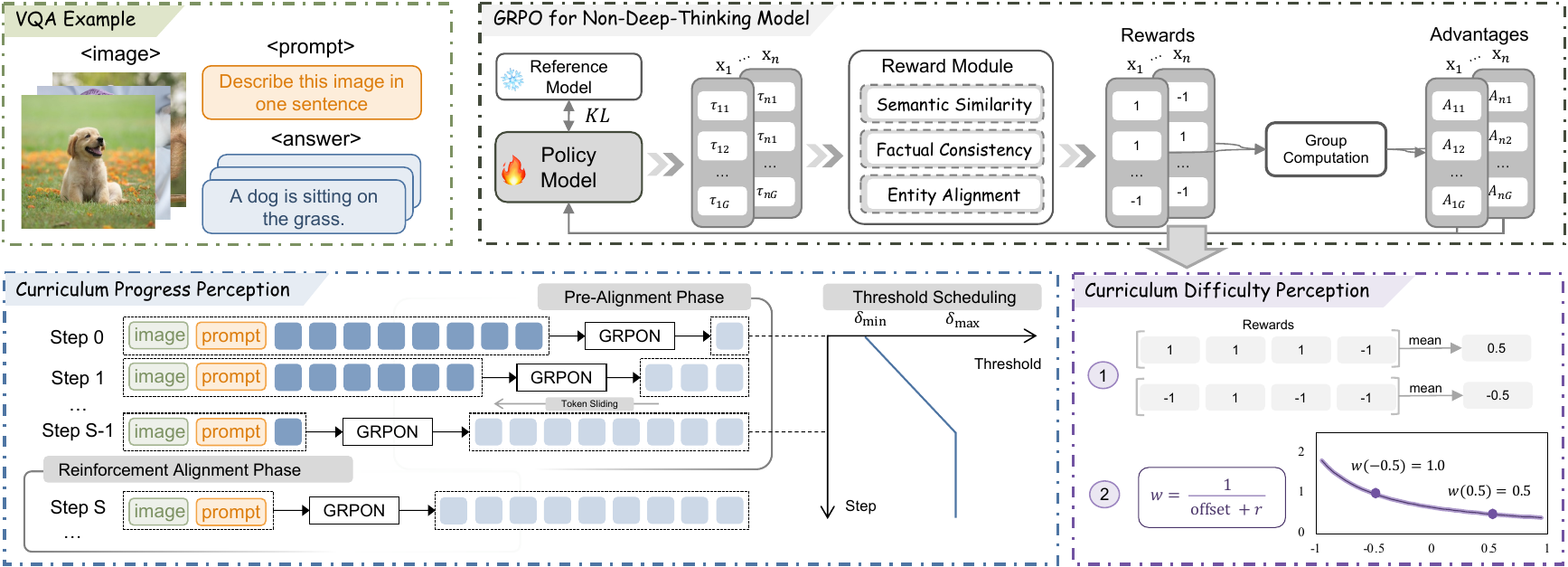}
    \caption{The Overview of RCPA. RCPA is a post-training framework that integrates domain knowledge acquisition with preference alignment in a curriculum-driven manner, built upon the GRPON (GRPO for Non-Deep-Thinking Models) framework for VQA-style tasks. It consists of two phases: Pre-Alignment, which introduces domain concepts with controlled constraints to bootstrap initial competence, and Reinforcement Alignment, which refines the model’s responses using full reward-driven optimization. Key components include Curriculum Progress Perception (CPP), which adjusts reward thresholds to match the model's evolving competence, and Curriculum Difficulty Perception (CDP), which prioritizes difficult samples to enhance training efficiency and prevent overfitting.}
    \label{fig:overview}
\end{figure*}

RCPA is a post-training paradigm that unifies domain knowledge acquisition with preference alignment under a curriculum-driven framework. As illustrated in Figure~\ref{fig:overview}, RCPA builds upon GRPON (GRPO for Non-Deep-Thinking Models) as the RL backbone for VQA-style tasks. A key design is progressive modulation mechanism, which decomposes adaptation into two coordinated phases: 

\textbf{Pre-Alignment Phase:} At early stages, the model lacks sufficient domain knowledge, RCPA employs partial output constraints to mitigate the challenges of sparse rewards and unavailable preference signals. By introducing domain concepts in a controlled manner (e.g., through partial answer exposure), this phase ensures valid response generation and bootstraps the model’s initial domain competence.

\textbf{Reinforcement Alignment Phase:} Once the model has attained a foundational understanding of the target domain, the training process gradually shifts toward full reward-driven optimization. Here, constraints are relaxed, and the model refines its responses based on reinforcement signals, achieving stronger preference alignment and higher task performance.

To enable smooth transitions, RCPA incorporates two curriculum-inspired modules: Curriculum Progress Perception (CPP) and Curriculum Difficulty Perception (CDP). We clarify the fundamental limitation of existing GRPO-based methods: their failure to effectively sample correct responses, leading to inadequate positive reward signals for meaningful learning. To address this, CPP is designed to gradually reduce the difficulty of sampling correct responses while refining target-domain knowledge acquisition. CPP dynamically schedules reward thresholds to match the model’s evolving competence—applying relatively low thresholds in the early stage to capture weak signals, and progressively increasing them to enforce stricter performance criteria as the model matures. CDP further enhances training efficiency by prioritizing difficult samples (with greater learning value) and down-weighting simpler ones, thereby preventing overfitting and encouraging high-value learning. Through this progressive and curriculum-driven approach, RCPA effectively coordinates deep knowledge integration with the preservation of existing capabilities, yielding a robust and adaptive alignment strategy.

\subsection{RL Backbone: GRPO for Non-Deep-Thinking Models}

We adapt the GRPO framework for VLMs that perform short-answer generation tasks (e.g., VQA), and denote this variant as GRPO for Non-Deep-Thinking (GRPON). A central component of GRPON is a rule-based reward function $R(o,y)$, which evaluates a generated output $o$ against a ground-truth answer $y$. The reward integrates three complementary criteria:

\textbf{Semantic Similarity ($S_s(o,y)$):} Measured as the cosine similarity between SentenceBERT~\citep{reimers-2019-sentence-bert} embeddings of $o$ and $y$.

\textbf{Factual Consistency ($S_f(o, y)$):} Assessed using a Natural Language Inference (NLI) model~\citep{DBLP:journals/corr/abs-1911-02116}, which checks bidirectional entailment and contradiction between $o$ and $y$.

\textbf{Entity Alignment ($S_e(o,y)$):} Computed as the F1-score over entities extracted by SpaCy~\citep{Honnibal_spaCy_Industrial-strength_Natural_2020} from $o$ and $y$.

The overall similarity score is defined as a weighted combination of these components:

\begin{small}
\begin{equation}
\label{eq:abpara}
    S(o, y) = \alpha \cdot S_s(o, y) + (1 - \alpha) \cdot \left[ \beta \cdot S_f(o, y) + (1 - \beta) \cdot S_e(o, y) \right]
\end{equation}
\end{small}
where $\alpha, \beta \in [0,1]$ are tunable hyperparameters that balance the contribution of each factor. This similarity score is then converted into a binary reward using a threshold $\delta$:
\begin{equation}
\label{eq:reward}
    R(o, y; \delta) = 
    \begin{cases} 
    +1 & \text{if}~S(o, y) > \delta, \\
    -1 & \text{otherwise}. 
    \end{cases} 
\end{equation}
Within our curriculum-driven framework, the threshold $\delta$ is scheduled dynamically across training steps: it begins at a relatively low value to enforce knowledge acquisition during the early stage, and is gradually increased to encourage precision and correctness as training progresses.

\subsection{Curriculum Progress Perception}

The Curriculum Progress Perception (CPP) module is designed to bootstrap learnable signals during the early stages of adaptation and to gradually transition toward free-form generation. It operates through two key mechanisms: (1) answer-prefix injection and (2) step-level threshold scheduling, with the overall goal of reducing the difficulty of sampling the correct answer and shaping an optimizable output distribution when the model initially lacks domain-specific knowledge.

\textbf{Answer-Prefix Injection.} For a training sample $\mathbf{x}_i = \{\text{image}_i, \text{prompt}_i\}$ with ground-truth answer $y_i = (y_{i,1}, ..., y_{i,|y_i|})$, let $s$ denote the current training step. CPP injects a prefix of length $k(s) =max(0, (1- \frac{s}{S} \times \sigma)) \cdot |y_i|$ into the input context, where $\sigma$ is the sliding token ratio. Each sample is treated as one training step, so $\sigma$ controls the overall number of steps in the Pre-Alignment Phase. The resulting context at step $s$ becomes $C(i,s) = (\text{image}_{i}, \text{prompt}_{i}, y_{i,\le k(s)})$, and the model is required to generate only the suffix $y_{i,>k(s)}$. At $s=0$, the model predicts only the EOS token, minimizing risk. At $s \ge S/\sigma$, no prefix is provided, reducing the task to standard full-response generation. This injected prefix acts as an explicit supervision anchor that strengthens cross-modal attention between the image and the partial textual answer, while also increasing the density of rewardable samples by reducing invalid completions, particularly crucial during early adaptation.

\textbf{Step-Level Threshold Scheduling.} 
To enhance progressive generation, CPP regulates the reward threshold $\delta$ (Equation~\ref{eq:reward}). In the early stages of training, when the model is exposed to simpler tasks, a lower threshold $\delta_{\text{min}}$ is used, which encourages exploration and allows the model to generate more diverse outputs so as to better acquire domain knowledge. As training progresses and the model gains more domain knowledge, the threshold is gradually increased to $\delta_{\text{max}}$, enforcing stricter alignment with the injected prefix to ensure accurate generation. This progression ensures that the model moves from broad exploration to focused, precise output generation. The threshold schedule follows a linear progression, with $\delta (s)$ formalized as:

\begin{equation}
    \delta (s) = \delta_{\text{min}} + (\delta _{\text{max}} - \delta _{\text{min}}) \times min(1,\frac{s}{S}\times \sigma),
\label{eq: threshold}
\end{equation}
\noindent where $\delta_{\max}, \delta_{\min}$ are predefined clipping parameters. The GRPON objective is applied only to suffix tokens ($t > k(s)$), conditioned on the context $C(s)$ and step-dependent threshold $\delta(s)$.

\subsection{Curriculum Difficulty Perception}
The Curriculum Difficulty Perception (CDP) module dynamically reweights training samples based on their difficulty, as reflected in the model’s real-time learning state. By prioritizing challenging samples and down-weighting simpler ones, CDP ensures that training resources are allocated to examples with the greatest learning value, thereby improving efficiency and mitigating overfitting.

For a given query, let $\mathbf{r} = (r_1, \dots, r_G)$ denote the rewards obtained from $G$ generated responses. The mean reward $\overline{r} = \text{mean}(\mathbf{r})$ serves as a proxy for sample difficulty: higher values indicate that the sample is easy, while lower values indicate greater difficulty. The sample weight $w$ is defined as $w = \frac{1}{\text{offset} + \bar{r}}$.

This weight is incorporated into the GRPON objective by scaling the advantage term $\hat{A}_{i,t}$ for each sample, ensuring that the model allocates greater capacity to high-value learning opportunities. We set the $\text{offset}$ to 1.5 to scale $w$ into a reasonable range, thereby stabilizing the training process. At training step $s$, the complete RCPA objective is expressed as:

\begin{equation}
\label{eq:RCPA_full}
\begin{aligned}
\mathcal{J}_{\text{RCPA}}&(\theta; s) = \mathbb{E}_{x, \{o_i\} \sim \pi_{\theta_{\text{old}}}} \biggl[ \frac{1}{G} \sum_{i, t} \frac{1}{\Delta_i} \min \Bigl( \rho_{i,t}(\theta) \hat{A}_{i,t}, \\
& \text{clip}(\rho_{i,t}, 1\!-\!\epsilon, 1\!+\!\epsilon) \hat{A}_{i,t} \Bigr) w - \gamma \mathbb{D}_{\text{KL}} \bigl[ \pi_{\theta} \| \pi_{\text{ref}} \bigr] \biggr], \\
& \text{where } \Delta_i = |o_i| - k(s), \sum_{i, t} = \sum_{i=1}^G \sum_{t=k(s)+1}^{|o_i|}.
\end{aligned}
\end{equation}

\noindent where the advantage $\hat{A}_{i,t}$ is computed using the reward $R(o_i, y; \delta(s))$. The overall training procedure is summarized in Algorithm~\ref{alg:RCPA}.
\input{TAB/alg}

%% file: TAB/alg.tex
\begin{algorithm}[t]
\caption{Reinforced Curriculum Pre-Alignment (RCPA)}
\label{alg:RCPA}

\renewcommand{\algorithmicrequire}{\textbf{Input:}}
\renewcommand{\algorithmicensure}{\textbf{Output:}}

\begin{algorithmic}[1]

\REQUIRE Pre-trained VLM $\pi_{\theta_{\text{pre}}}$; Target dataset $\mathcal{D}_{\text{target}} = \{(\mathbf{x}_i, y_i)\}_{i=1}^N$; Curriculum parameters $(S, \delta_{\max}, \delta_{\min}, \sigma)$; GRPO hyperparameters $(\alpha, \beta, \epsilon, G)$.
\ENSURE Domain-adapted model $\pi_{\theta}$.

\STATE Initialize $\pi_{\theta} \leftarrow \pi_{\theta_{\text{pre}}}$, reference model $\pi_{\theta_{\text{ref}}} \leftarrow \pi_{\theta_{\text{pre}}}$, step $s \leftarrow 0$, $S \leftarrow N$
% \FOR{step $s = 0, 1, \dots, S$} 
    \STATE Sample minibatch $B \subset \mathcal{D}_{\text{target}}$
    \FOR{sample $(\mathbf{x}, y) \in B$}
        \STATE Compute prefix length $k(s) = max(0,(1 - \frac{s}{S}\times \sigma)) \cdot |y|$ 
        \STATE Construct context $C(s) = (\mathbf{x}, y_{\le k(s)})$ 
        \STATE Generate $G$ candidate outputs$\{o_i\}_{i=1}^G \sim \pi_{\theta}(\cdot | C(s))$ 
        \FORALL{$o_i \in \{o_i\}_{i=1}^G $}
            \STATE Compute reward $R(o_i, y; \delta(s))$ using Eq.~\ref{eq:reward} 
            \STATE Compute mean reward $\bar{r} = \text{mean}(\{R(o_i, y; \delta(s))\}_{i=1}^G)$ 
            \STATE Compute difficulty weight $w = \frac{1}{\text{offset} + \bar{r}}$  
            \STATE Compute curriculum threshold $\delta(s) = \delta_{\min} + (\delta_{\max} - \delta_{\min}) \times min(1, \tfrac{s}{S} \times \sigma)$ 
            \STATE Compute RCPA objective $\mathcal{J}_{\text{RCPA}}(\theta; s)$ using Eq.~\ref{eq:RCPA_full} 
            \STATE Update $\pi_{\theta}$ by gradient ascent on $\mathcal{J}_{\text{RCPA}}(\theta; s)$
        \ENDFOR
        \STATE s = s+1
    \ENDFOR
% \ENDFOR
\RETURN $\pi_{\theta}$
\end{algorithmic}
\end{algorithm}

%% file: TEX/5.Experiment.tex
\section{Experiment}
We begin this section by outlining our experimental setup (Section~\ref{sec: setup}). We then present a comprehensive comparison of RCPA against current state-of-the-art methods (Section~\ref{sec: comparison}). This is followed by ablation studies that assess the contribution of each core component (Section~\ref{sec: ablation_study}) and parameter sensitivity analyses that evaluate the robustness of key hyperparameters (Section~\ref{sec: para_study}). Additional results on computational efficiency, the effectiveness of cold start in RL, optimization stability, and generalization are provided in Section~\ref{sec: com}, Section~\ref{sec: cold}, and Section~\ref{sec: osg}.

\subsection{Experimental Setup}
\label{sec: setup}
\textbf{Benchmark Datasets.}
To evaluate RCPA’s adaptability and performance, we conduct experiments on three benchmark datasets spanning diverse domains: image captioning, geometric problem-solving, and medical X-ray diagnostics.
\begin{itemize}[leftmargin=1em]
 \item COCO Caption~\citep{chen2015microsoft}: A widely used dataset for image captioning, containing 123,287 images. We use the original training and test splits of the dataset.
\item Geo170K~\citep{gao2025gllava}: A dataset for geometric problem-solving. It is divided into Phase 1 (non-deep thinking, direct-answer tasks) and Phase 2 (deep thinking, multi-step tasks). We use Phase 1, which contains 60,252 samples, with 57,252 for training and 3,000 for testing.
\item OpenI~\citep{demner2012design}: A chest X-ray diagnostic dataset with 6,423 images and corresponding radiological reports. The task is to generate concise and clinically accurate diagnostic descriptions directly from X-ray images. The training set consists of 5,423 images, while the test set includes 1,000 images.
\end{itemize}

These datasets enable us to evaluate RCPA’s performance across both general domain (e.g., COCO Caption) and specific domain (e.g., Geo170K, OpenI) tasks, providing a comprehensive assessment of domain-adaptive capabilities.
\input{TAB/all}

\textbf{Baselines.} We employ Qwen2.5-VL-7B \citep{Qwen2.5-VL} as the base model and compare the following adaptation strategies. (1) \textbf{BASE}: Direct inference using the pre-trained Qwen2.5-VL-7B model. (2) \textbf{SFT-based Methods}: Including Parameter-Efficient Fine-tuning (PEFT) via LoRA~\citep{Hu2021LoRALA} and Full Fine-tuning (FFT). Furthermore, drawing on relevant methods in incremental learning ~\citep{Kirkpatrick2016OvercomingCF}, we incorporate the Kullback-Leibler (KL) divergence loss into Full Fine-Tuning (FFT), and define this improved approach as Continual Full Fine-Tuning (CFFT). Specifically, this KL divergence loss imposes a constraint that aligns the output distribution of the fine-tuned model with that of the pre-trained model, effectively mitigating the overwriting of general capabilities by domain-specific signals.
(3) \textbf{RL-based Methods}: Group Relative Policy Optimization (GRPO) \citep{Shao2024DeepSeekMathPT}, Decoupled Clip and Dynamic sAmpling Policy Optimization (DAPO) \citep{Yu2025DAPOAO}, GRPO for Non-Deep-Thinking Models (GRPON). Cold start is added to GRPON to evaluate our method comprehensively.

\textbf{Evaluation Metrics.} We evaluate RCPA using a combination of task-specific metrics for domain knowledge injection and generalization metrics for preserving general-purpose performance. Domain-specific ability is assessed with \textbf{BLEU-1} \citep{papineni-etal-2002-bleu}, \textbf{CIDEr} \citep{Vedantam2014CIDErCI}, \textbf{ROUGE-L} \citep{lin2004rouge}, and \textbf{SPICE} \citep{anderson2016spice} to measure n-gram overlap, semantic consistency, long-sequence similarity, and structural alignment. General-purpose capabilities are evaluated using \textbf{MMMU} \citep{Fu2023MMEAC} for cross-domain reasoning, \textbf{MME} \citep{Fu2023MMEAC} for multimodal understanding, and \textbf{IFEval} \citep{Zhou2023InstructionFollowingEF}—including IFEval-Prompt (IFEval-P) and  IFEval-Instruct (IFEval-I)—to measure instruction-following fidelity and consistency with human intent.

\textbf{Implementation Details.} For model training and inference, we set the hyperparameters $\alpha$ and $\beta$ in Equation~\ref{eq:abpara} to 0.6 and 0.7, respectively. In Equation~\ref{eq: threshold}, the threshold parameters are configured with $\delta_{\text{max}} = 0.7$, $\delta_{\text{min}} = 0.8$, and $\sigma = 16$. For the RCPA objective (Equation~\ref{eq:RCPA_full}), the regularization coefficient $\gamma$ is set to 0.01. All experiments are implemented using the EasyR1 RL framework \citep{zheng2025easyr1} and LlamaFactory SFT framework \citep{zheng2024llamafactory} for VLMs and conducted on a Linux-based server equipped with NVIDIA GPUs. 

\vspace{-0.2cm}
\subsection{Performance Comparison}
\label{sec: comparison}
Main experimental results are summarized in Table~\ref{tab: comparison}. Across all benchmarks, RCPA delivers competitive domain-specific performance while preserving general-purpose capabilities. Analysis of the COCO Caption dataset reveals several key insights:
\begin{itemize}[leftmargin=1em]

\vspace{-0.2cm}
\item FFT vs. Generality: While FFT achieves peak domain scores (e.g., 1.0172 CIDEr), it induces severe degradation in general benchmarks, with declines of 8.78\% on MMMU, 1598.26 on MME, and over 36\% on IFEval-I.

\item PEFT Limitations: LoRA mitigates generality loss but compromises instruction-following—dropping to 0.5416 on IFEval-P—and remains less effective at domain knowledge acquisition, scoring only 0.0862 on CIDEr.

\item GRPON Efficacy: GRPON surpasses standard GRPO in domain-specific captioning, yielding significant gains in BLEU-1 (+21.92\%), ROUGE-L (+8.89\%), CIDEr (+4.9\%), and SPICE (+4.93\%). This stems from bypassing unnecessary deep-thinking for non-reasoning tasks.

\item RCPA Superiority: RCPA matches FFT’s domain performance (e.g., 0.7478 BLEU-1 vs. FFT's 0.7581) while maintaining high general capability scores, such as 0.7326 on IFEval-I, striking a superior balance between specialization and generalization.
\end{itemize}
Consistent trends are observed on Geo170K and OpenI:
\textbf{Geo170K}: RCPA achieves a 2.2821 CIDEr, comparable to FFT's 2.3109, while significantly outperforming FFT in general understanding (MMMU: 0.5122 vs. 0.4667).
\textbf{OpenI}: In medical diagnostics, RCPA maintains a 0.3342 BLEU-1 and 0.2325 ROUGE-L, while preserving a 0.7062 IFEval-I score, far exceeding FFT’s 0.6367. 

\begin{figure*}
    \centering
    \vspace{-0.2cm}
    \includegraphics[width=0.95\linewidth]{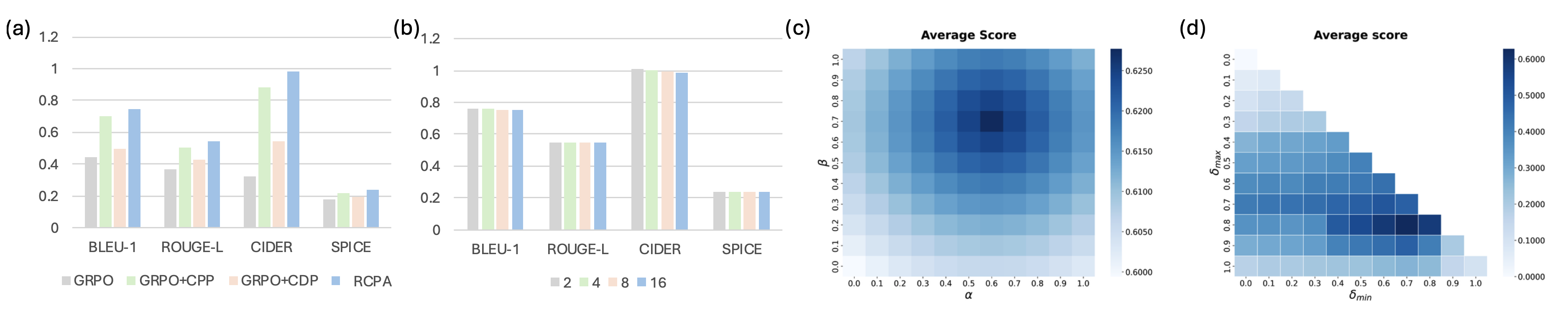}
    \caption{Results of Ablation and Parameter studies. (a) Ablation on COCO Captions demonstrates the contribution of CPP and CDP, with CPP leading to larger gains in domain-specific learning, while CDP enhances specialization by reweighting samples based on difficulty. (b) The impact of the sliding token ratio ($\sigma$) on domain knowledge learning and training efficiency reveals the optimal value of 16. (c) and (d) Parameter study results show that optimal performance is achieved with $\alpha=0.6, \beta=0.7, \delta_{\text{min}} = 0.7$, and $\delta_{\text{max}}=0.8$, based on aggregated evaluation metrics.}
    \label{fig:allablation}
    \vspace{-0.2cm}
\end{figure*}

\subsection{Ablation Study}
\label{sec: ablation_study}
Ablation experiments on COCO Captions (Figure~\ref{fig:allablation} (a)) confirm that both CPP and CDP enhance domain-specific learning, with CPP providing more substantial gains. By injecting and progressively shortening answer prefixes, CPP decomposes response generation into staged subtasks. This approach lowers early optimization barriers, increases reward signal density, and strengthens cross-modal grounding. Conversely, CDP improves specialization by dynamically prioritizing difficult, underlearned samples over easy, overlearned cases. This mechanism prevents overfitting while encouraging broader coverage of the target domain. Together, they form a complementary curriculum: CPP guides the learning path through reward densification, while CDP sharpens the focus via difficulty-aware prioritization.

\subsection{Parameter Study}
\label{sec: para_study}
\textbf{Impact of sliding token ratio.} Figure~\ref{fig:allablation} (b) shows that a larger $\sigma$ accelerates answer token restoration. While this reduces training iterations, it undermines domain-specific learning by providing insufficient time for the model to thoroughly absorb and align with specialized knowledge, leading to superficial mastery. Conversely, a smaller ratio facilitates deeper knowledge acquisition but reduces training efficiency. To balance integration depth with practical efficiency, we empirically select $\sigma=16$. This setting moderates token recovery to support incremental learning while remaining computationally feasible.

\textbf{Influence of weight and threshold parameters:} We assess the impact of $\alpha$, $\beta$, $\delta_{min}$, and $\delta_{max}$ using an aggregate score of BLEU-1, CIDEr, ROUGE-L, and SPICE to capture diverse aspects of generation quality. For each parameter, we fix one value and vary others from 0 to 1 in steps of 0.1. Results indicate that $\alpha=0.6$ and $\beta=0.7$ achieve the optimal balance between semantic similarity, factual consistency, and entity alignment. Similarly, the most effective thresholds for progressive generation are $\delta_{min}=0.7$ and $\delta_{max}=0.8$. These configurations are adopted for all subsequent experiments.

\subsection{Computation Efficiency}
\label{sec: com}

\begin{figure*}
    \centering
    \includegraphics[width=0.99\linewidth]{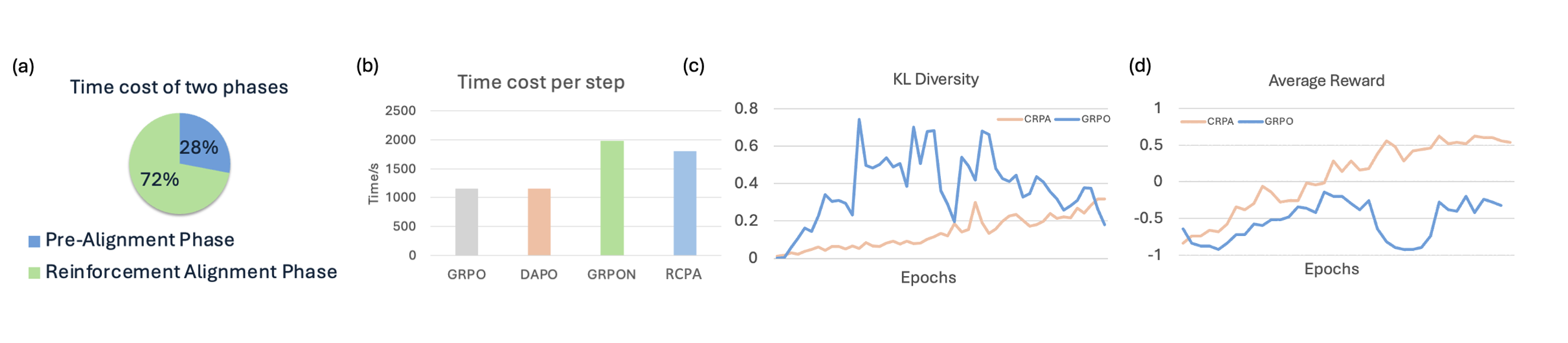}
    \caption{(a) The pre-alignment stage accounts for 28\% of the total training time. (b) Under the same batch size, RCPA increases the computation time per step by 56\% compared with GRPO. (c) Compared with GRPO, our RCPA reduces policy update variance by about 41\% (measured via KL divergence between consecutive policies). (d) Compared with GRPO, our proposed RCPA enables a continuous and stable increment in reward values throughout the entire training process, demonstrating more favorable reward growth characteristics in terms of both sustainability and stability.}
    \label{fig:rl_reward}
\end{figure*}

As shown in Figure~\ref{fig:rl_reward} (a) and (b), the first pre-alignment stage accounts for 28\% of the total training time. Compared with GRPO, RCPA is designed with a more sophisticated value function, which consequently increases the computation time per step by 56\% under the same batch size. Nevertheless, considering the significant performance gains brought by our method, such computational overhead is well-justified.

\input{TAB/cold}

\subsection{The Effectiveness of Cold Start in RL}
\label{sec: cold}

In order to validate the efficacy of cold-start strategies, we conduct controlled experiments by performing cold-start with partial target-domain training data, specifically adopting 0.1 epoch and 1.0 epoch of data for the cold-start phase, respectively. As shown in Table~\ref{tab: coldstart}, the experimental results reveal two key findings:
(1) Even a modest cold-start (0.1 epoch) leads to improved target-domain performance, yet it irreversibly degrades the model’s general capabilities—a degradation that cannot be recuperated via subsequent reinforcement learning (RL) fine-tuning.
(2) Increasing the cold-start data volume to 1.0 epoch yields marginally superior target-domain performance compared to our RCPA method, but this gain comes at the expense of substantial generalization degradation (e.g., the MMMU score drops from 0.5122 to 0.4522 on the COCO Caption dataset).

\subsection{Optimization Stability and Generalization}
\label{sec: osg}

We visualized the Kullback-Leibler (KL) divergence and average reward throughout the training process. As illustrated in Figure~\ref{fig:rl_reward} (c) and (d), our RCPA method enables stable training: its KL divergence changes in a relatively smooth pattern, and the overall KL divergence is reduced by 41\% compared to that of GRPO. Meanwhile, the visualization of the average reward reveals that, benefiting from its targeted design, the average reward of our method steadily rises to a relatively high level—this indicates that our method can effectively learn the knowledge of the target domain. In contrast, GRPO exhibits severe fluctuations in the average reward, and its overall average reward does not increase significantly. All these observations collectively demonstrate the superiority of our proposed method.

%% file: TAB/all.tex
% \usepackage{multirow}
\begin{table*}[t]
\centering
\resizebox{0.95\textwidth}{!}{
\begin{tabular}{clcccccccc}
\toprule
\multirow{2}{*}{Datasets} & \multicolumn{1}{c}{\multirow{2}{*}{Methods}} & \multicolumn{4}{c}{Domain-Specific Ability} & \multicolumn{4}{c}{General-Purpose Ability}\\ \cmidrule(r){3-6}\cmidrule(r){7-10} 
& \multicolumn{1}{c}{} & BLEU-1& ROUGE-L & CIDEr & SPICE & MMMU & MME & IFEval-P & IFEval-I \\ \midrule
\multirow{7}{*}{\shortstack{COCO\\Caption}} 
& BASE & 0.4457& 0.3672& 0.2259 & 0.1783& 0.5122 & \underline{2333.36} & 0.6211& 0.7038\\ \cmidrule{2-10} 
& PEFT & 0.3722& 0.3081& 0.1231 & 0.0862 & \textbf{0.6067} & \textbf{2448.67} & 0.5416& 0.6535\\
& FFT& \textbf{0.7581}& \textbf{0.5474}& \textbf{1.0172} & \textbf{0.2385}& 0.4244 & 735.10 & 0.2070& 0.3405\\ 
% EWC}& 0.6347& 0.4776& 0.7234 & 0.2001& 0.4911 & 1921.30 & 0.5164& 0.6233\\
&CFFT& 0.6518 & 0.4824 & 0.7654 & 0.2034 & 0.4967 & 1934.32 & 0.5324 & 0.6419 \\ \cmidrule{2-10} 
& GRPO & 0.2245& 0.2767& 0.2699 & 0.1297& 0.5222 & 2301.53 & \underline{0.6506}& \underline{0.7410}\\
& DAPO & 0.2431& 0.2798& 0.2687 & 0.1301& 0.5111 & 2330.27 & \textbf{0.6577}& \textbf{0.7423}\\
& GRPON& 0.4437& 0.3656& 0.3189 & 0.1790 & 0.5100 & 2315.97 & 0.6299& 0.7238\\
% GRPON+CS(0.1)}& 0.6234& 0.4326& 0.6776 & 0.1997 & 0.4811 & 1976.37 & 0.5267& 0.5324\\
% GRPON+CS(1.0)}& 0.7501& 0.5444& 0.9921 & 0.2383 & 0.4522 & 901.43 & 0.3012& 0.4123\\
& \cellcolor{gray!30}{RCPA} & \cellcolor{gray!30}{\underline{0.7478}}& \cellcolor{gray!30}{\underline{0.5432}}& \cellcolor{gray!30}{\underline{0.9814}} & \cellcolor{gray!30}{\underline{0.2383}}& \cellcolor{gray!30}{\underline{0.5278}} & \cellcolor{gray!30}{2289.18} & \cellcolor{gray!30}{0.6470}& \cellcolor{gray!30}{0.7326}\\ \hline \hline
\multirow{7}{*}{Geo170K}
& BASE & 0.3859& 0.3014& 0.2740 & 0.2901& 0.5122 & 2333.36 & 0.6211& 0.7038\\ \cmidrule{2-10} 
& PEFT & 0.0901& 0.1192& 0.0009 & - &  \textbf{0.6067} &  \textbf{2449.67} & 0.5360& 0.6535\\
& FFT&  \textbf{0.6098}&  \textbf{0.5526}&  \textbf{2.3109} &  \textbf{0.5627}& 0.4667 & 2172.37 & 0.5693& 0.6451\\
% EWC}& 0.5587& 0.4976& 1.7801 & 0.4987& 0.5011 & 2176.47 & 0.5761& 0.6589\\
&CFFT& 0.5719& 0.5091& 1.8827 & 0.5079& 0.5078 & 2199.59 & 0.5888 & 0.6676 \\ \cmidrule{2-10} 
& GRPO & 0.3799& 0.3113& 0.2661 & 0.2878& 0.5022 & \underline{2346.08} & 0.6373& \underline{0.7131}\\
& DAPO & 0.3835& 0.3189& 0.2776 & 0.2989& 0.5111 & 2319.25 & 0.6285& 0.7110\\
& GRPON& 0.4086& 0.3431& 0.3543 & 0.3350 & 0.5122 & 2320.54 & \underline{0.6414}& 0.7062\\
% GRPON+CS(0.1)}& 0.5545& 0.4943& 1.6893 & 0.4923 & 0.4922 & 2219.54 & 0.6026& 0.7062\\
% GRPON+CS(1.0)}& 0.6012& 0.5504& 2.2896 & 0.5623 & 0.4767 & 2199.34 & 0.5801& 0.6594\\
& \cellcolor{gray!30}{RCPA} & \cellcolor{gray!30}{\underline{0.5998}}& \cellcolor{gray!30}{\underline{0.5501}}& \cellcolor{gray!30}{\underline{2.2821}} & \cellcolor{gray!30}{\underline{0.5623}}& \cellcolor{gray!30}{\underline{0.5122}} & \cellcolor{gray!30}{2315.37} & \cellcolor{gray!30}{ \textbf{0.6414}}& \cellcolor{gray!30}{ \textbf{0.7278}}\\ \hline \hline
\multirow{7}{*}{OpenI}& 
BASE & 0.1155& 0.1299& 0.0002 & 0.0988& 0.5122 & 2333.36 & 0.6211& 0.7038\\ \cmidrule{2-10} 
& PEFT & 0.0786& 0.0977& -& 0.0871& \textbf{0.6067} & \textbf{2449.67} & 0.5508& 0.6631\\
& FFT& \textbf{0.3396}& \textbf{0.2399}& \textbf{0.0903} & \textbf{0.1900}& 0.4111 & 1623.35 & 0.5323& 0.6367\\ 
% EWC}& 0.2647& 0.1861& 0.0801 & 0.1652& 0.4667 & 2089.34 & 0.5368& 0.6567\\

&CFFT& 0.2698& 0.1889& 0.0813 & 0.1698& 0.4711 & 2100.32& 0.5578& 0.6719\\ \cmidrule{2-10} 
& GRPO & 0.1179& 0.1309& 0.0003 & 0.0994 & \underline{0.5122} & \underline{2356.23} &0.6248& 0.7062\\
& DAPO & 0.1165& 0.1356& 0.0003 & 0.0998 & 0.5111 & 2334.65 & \underline{0.6267}& \underline{0.7098}\\
& GRPON& 0.1182& 0.1311& 0.0003 & 0.0999& 0.5044 & 2315.37 & 0.6192& \textbf{0.7110}\\
% GRPON+CS(0.1)}& 0.2723& 0.1820& 0.0823 & 0.1379& 0.4743 &1989.34 & 0.5792& 0.6723\\
% GRPON+CS(1.0)}& 0.3378& 0.2311& 0.0896 & 0.1839& 0.4311 & 1710.69 & 0.5511& 0.6501\\
& \cellcolor{gray!30}{RCPA} & \cellcolor{gray!30}{\underline{0.3342}}& \cellcolor{gray!30}{\underline{0.2325}}& \cellcolor{gray!30}{\underline{0.0886}} & \cellcolor{gray!30}{\underline{0.1814}}& \cellcolor{gray!30}{0.5011} & \cellcolor{gray!30}{2321.40} & \cellcolor{gray!30}{\textbf{0.6285}}& \cellcolor{gray!30}{0.7062}\\ \bottomrule
\end{tabular}
}

\caption{Performance comparison of different recommendation methods in terms of Domain-Specific Ability and General Ability. COCO Caption, Geo170k, and OpenI are used as benchmark datasets. The results across these datasets demonstrate that RCPA achieves domain-specific performance comparable to that of FFT, while also preserving general capabilities. `-' indicates that the metric is either too small or the generated output is too long, causing it to be truncated and thus unable to be calculated. The best results are shown in bold. The second-best results are marked with an underline.}
\label{tab: comparison}
\end{table*}
\vspace{-0.2cm}
% The best results are indicated in bold text, and the second-best results are underlined.

%% file: TAB/cold.tex
\begin{table*}[]
\centering
\resizebox{0.95\textwidth}{!}{
\begin{tabular}{clcccccccc}
\toprule
\multirow{2}{*}{Datasets} & \multicolumn{1}{c}{\multirow{2}{*}{Methods}} & \multicolumn{4}{c}{Domain-Specific Ability} & \multicolumn{4}{c}{General-Purpose Ability}\\ \cmidrule(r){3-6}\cmidrule(r){7-10} 
& \multicolumn{1}{c}{} & BLEU-1& ROUGE-L & CIDEr & SPICE & MMMU & MME & IFEval-P & IFEval-I \\ \midrule
\multirow{7}{*}{\shortstack{COCO\\Caption}} 
& BASE & 0.4457& 0.3672& 0.2259 & 0.1783& \underline{0.5122} & \textbf{2333.36} & 0.6211& 0.7038\\ \cmidrule{2-10} 
& GRPON& 0.4437& 0.3656& 0.3189 & 0.1790 & 0.5100 & \underline{2315.97} & \underline{0.6299}& \underline{0.7238}\\
& GRPON+CS(0.1)& 0.6234 & 0.4326 & 0.6776 & 0.1997 & 0.4811 & 1976.37 & 0.5267 & 0.5324\\
& GRPON+CS(1.0)& \textbf{0.7501} & \textbf{0.5444} & \textbf{0.9921} & \textbf{0.2396}& 0.4522 & 901.43 & 0.3012 & 0.4123 \\
& \cellcolor{gray!30}{RCPA} & \cellcolor{gray!30}{\underline{0.7478}}& \cellcolor{gray!30}{\underline{0.5432}}& \cellcolor{gray!30}{\underline{0.9814}} & \cellcolor{gray!30}{\underline{0.2383}}& \cellcolor{gray!30}{\textbf{0.5278}} & \cellcolor{gray!30}{2289.18} & \cellcolor{gray!30}{\textbf{0.6470}}& \cellcolor{gray!30}{\textbf{0.7326}}\\ \hline \hline
\multirow{7}{*}{Geo170K}
& BASE & 0.3859& 0.3014& 0.2740 & 0.2901& 0.5122 & \textbf{2333.36} & 0.6211& 0.7038\\ \cmidrule{2-10} 
& GRPON& 0.4086& 0.3431& 0.3543 & 0.3350 & \underline{0.5122} & \underline{2320.54} & \underline{0.6414}& \underline{0.7062}\\
& GRPON+CS(0.1)& 0.5545 & 0.4943 & 1.6893 & 0.4923 & 0.4922 & 2219.54 & 0.6026 & 0.7062\\
& GRPON+CS(1.0)& \textbf{0.6012} & \textbf{0.5504} & \textbf{2.2896} & \textbf{0.5629} & 0.4767 & 2199.34 & 0.5801 & 0.6594 \\
& \cellcolor{gray!30}{RCPA} & \cellcolor{gray!30}{\underline{0.5998}}& \cellcolor{gray!30}{\underline{0.5501}}& \cellcolor{gray!30}{\underline{2.2821}} & \cellcolor{gray!30}{\underline{0.5623}}& \cellcolor{gray!30}{\textbf{0.5122}} & \cellcolor{gray!30}{2315.37} & \cellcolor{gray!30}{ \textbf{0.6414}}& \cellcolor{gray!30}{ \textbf{0.7278}}\\ \hline \hline
\multirow{7}{*}{OpenI}& 
BASE & 0.1155& 0.1299& 0.0002 & 0.0988& \textbf{0.5122} & \textbf{2333.36} & \underline{0.6211}& 0.7038\\ \cmidrule{2-10} 
& GRPON& 0.1182& 0.1311& 0.0003 & 0.0999& \underline{0.5044} & 2315.37 & 0.6192& \textbf{0.7110}\\
& GRPON+CS(0.1)& 0.2723& 0.1820& 0.0823 & 0.1379& 0.4743 &1989.34 & 0.5792& 0.6723\\
& GRPON+CS(1.0)& \textbf{0.3378}& \underline{0.2311}& \textbf{0.0896} & \textbf{0.1839}& 0.4311 & 1710.69 & 0.5511& 0.6501\\
& \cellcolor{gray!30}{RCPA} & \cellcolor{gray!30}{\underline{0.3342}}& \cellcolor{gray!30}{\textbf{0.2325}}& \cellcolor{gray!30}{\underline{0.0886}} & \cellcolor{gray!30}{\underline{0.1814}}& \cellcolor{gray!30}{0.5011} & \cellcolor{gray!30}{\underline{2321.40}} & \cellcolor{gray!30}{\textbf{0.6285}}& \cellcolor{gray!30}{\underline{0.7062}}\\ \bottomrule
\end{tabular}
}
\caption{Performance comparison of different cold start data volumes in terms of Domain-Specific Ability and General Ability. COCO Caption, Geo170k, and OpenI are used as benchmark datasets. CS(*) means that data of * epoch is used in the cold start. The best results are shown in bold. The second-best results are marked with an underline.}
\label{tab: coldstart}
\end{table*}

%% file: TEX/6.Conclusion.tex
\section{Conclusion}
We propose Reinforced Curriculum Pre-Alignment (RCPA), a novel framework that enables VLMs to acquire specialized domain knowledge without compromising general-purpose capabilities. By integrating a two-phase process: pre-alignment for stable knowledge grounding and reinforcement alignment for behavioral refinement, RCPA effectively bridges the gap between SFT and RL-based methods. The inclusion of CPP and CDP modules further ensures efficient learning while preventing optimization collapse and overfitting. Extensive experiments across medical and geometric benchmarks demonstrate that RCPA achieves domain-specific performance competitive with FFT while significantly outperforming existing baselines in retaining general multimodal understanding and instruction-following.

%% file: reference.bib
@article{wang2025internvl3_5,
  title={InternVL3.5: Advancing Open-Source Multimodal Models in Versatility, Reasoning, and Efficiency},
  author={Wang, Weiyun and Gao, Zhangwei and Gu, Lixin and Pu, Hengjun and Cui, Long and Wei, Xingguang and Liu, Zhaoyang and Jing, Linglin and Ye, Shenglong and Shao, Jie and others},
  journal={arXiv preprint arXiv:2508.18265},
  year={2025}
}

@article{zhu2025internvl3,
  title={Internvl3: Exploring advanced training and test-time recipes for open-source multimodal models},
  author={Zhu, Jinguo and Wang, Weiyun and Chen, Zhe and Liu, Zhaoyang and Ye, Shenglong and Gu, Lixin and Tian, Hao and Duan, Yuchen and Su, Weijie and Shao, Jie and others},
  journal={arXiv preprint arXiv:2504.10479},
  year={2025}
}

@article{chen2024expanding,
  title={Expanding Performance Boundaries of Open-Source Multimodal Models with Model, Data, and Test-Time Scaling},
  author={Chen, Zhe and Wang, Weiyun and Cao, Yue and Liu, Yangzhou and Gao, Zhangwei and Cui, Erfei and Zhu, Jinguo and Ye, Shenglong and Tian, Hao and Liu, Zhaoyang and others},
  journal={arXiv preprint arXiv:2412.05271},
  year={2024}
}

@article{wang2024mpo,
  title={Enhancing the Reasoning Ability of Multimodal Large Language Models via Mixed Preference Optimization},
  author={Wang, Weiyun and Chen, Zhe and Wang, Wenhai and Cao, Yue and Liu, Yangzhou and Gao, Zhangwei and Zhu, Jinguo and Zhu, Xizhou and Lu, Lewei and Qiao, Yu and Dai, Jifeng},
  journal={arXiv preprint arXiv:2411.10442},
  year={2024}
}

@article{gao2024mini,
  title={Mini-InternVL: a flexible-transfer pocket multi-modal model with 5\% parameters and 90\% performance},
  author={Gao, Zhangwei and Chen, Zhe and Cui, Erfei and Ren, Yiming and Wang, Weiyun and Zhu, Jinguo and Tian, Hao and Ye, Shenglong and He, Junjun and Zhu, Xizhou and others},
  journal={Visual Intelligence},
  volume={2},
  number={1},
  pages={1--17},
  year={2024},
  publisher={Springer}
}

@article{Zhou2023InstructionFollowingEF,
  title={Instruction-Following Evaluation for Large Language Models},
  author={Jeffrey Zhou and Tianjian Lu and Swaroop Mishra and Siddhartha Brahma and Sujoy Basu and Yi Luan and Denny Zhou and Le Hou},
  journal={ArXiv},
  year={2023},
  volume={abs/2311.07911},
  url={https://api.semanticscholar.org/CorpusID:265157752}
}

@article{Fu2023MMEAC,
  title={MME: A Comprehensive Evaluation Benchmark for Multimodal Large Language Models},
  author={Chaoyou Fu and Peixian Chen and Yunhang Shen and Yulei Qin and Mengdan Zhang and Xu Lin and Zhenyu Qiu and Wei Lin and Jinrui Yang and Xiawu Zheng and Ke Li and Xing Sun and Rongrong Ji},
  journal={ArXiv},
  year={2023},
  volume={abs/2306.13394},
  url={https://api.semanticscholar.org/CorpusID:259243928}
}

@misc{liu2024llavanext,
    title={LLaVA-NeXT: Improved reasoning, OCR, and world knowledge},
    url={https://llava-vl.github.io/blog/2024-01-30-llava-next/},
    author={Liu, Haotian and Li, Chunyuan and Li, Yuheng and Li, Bo and Zhang, Yuanhan and Shen, Sheng and Lee, Yong Jae},
    month={January},
    year={2024}
}

@misc{liu2023improvedllava,
      title={Improved Baselines with Visual Instruction Tuning}, 
      author={Liu, Haotian and Li, Chunyuan and Li, Yuheng and Lee, Yong Jae},
      publisher={arXiv:2310.03744},
      year={2023},
}

@article{Qwen2.5-VL,
  title={Qwen2.5-VL Technical Report},
  author={Bai, Shuai and Chen, Keqin and Liu, Xuejing and Wang, Jialin and Ge, Wenbin and Song, Sibo and Dang, Kai and Wang, Peng and Wang, Shijie and Tang, Jun and Zhong, Humen and Zhu, Yuanzhi and Yang, Mingkun and Li, Zhaohai and Wan, Jianqiang and Wang, Pengfei and Ding, Wei and Fu, Zheren and Xu, Yiheng and Ye, Jiabo and Zhang, Xi and Xie, Tianbao and Cheng, Zesen and Zhang, Hang and Yang, Zhibo and Xu, Haiyang and Lin, Junyang},
  journal={arXiv preprint arXiv:2502.13923},
  year={2025}
}

@article{Qwen2-VL,
  title={Qwen2-VL: Enhancing Vision-Language Model's Perception of the World at Any Resolution},
  author={Wang, Peng and Bai, Shuai and Tan, Sinan and Wang, Shijie and Fan, Zhihao and Bai, Jinze and Chen, Keqin and Liu, Xuejing and Wang, Jialin and Ge, Wenbin and Fan, Yang and Dang, Kai and Du, Mengfei and Ren, Xuancheng and Men, Rui and Liu, Dayiheng and Zhou, Chang and Zhou, Jingren and Lin, Junyang},
  journal={arXiv preprint arXiv:2409.12191},
  year={2024}
}

@article{Qwen-VL,
  title={Qwen-VL: A Versatile Vision-Language Model for Understanding, Localization, Text Reading, and Beyond},
  author={Bai, Jinze and Bai, Shuai and Yang, Shusheng and Wang, Shijie and Tan, Sinan and Wang, Peng and Lin, Junyang and Zhou, Chang and Zhou, Jingren},
  journal={arXiv preprint arXiv:2308.12966},
  year={2023}
}

@article{Gekhman2024DoesFL,
  title={Does Fine-Tuning LLMs on New Knowledge Encourage Hallucinations?},
  author={Zorik Gekhman and G. Yona and Roee Aharoni and Matan Eyal and Amir Feder and Roi Reichart and Jonathan Herzig},
  journal={ArXiv},
  year={2024},
  volume={abs/2405.05904},
  url={https://api.semanticscholar.org/CorpusID:269635770}
}

@article{Guo2023FinetuningSF,
  title={Fine-tuning Strategies for Domain Specific Question Answering under Low Annotation Budget Constraints},
  author={Kunpeng Guo and Dennis Diefenbach and Antoine Gourru and Christophe Gravier},
  journal={2023 IEEE 35th International Conference on Tools with Artificial Intelligence (ICTAI)},
  year={2023},
  pages={166-171},
  url={https://api.semanticscholar.org/CorpusID:252620041}
}

@misc{zheng2025easyr1,
  title        = {EasyR1: An Efficient, Scalable, Multi-Modality RL Training Framework},
  author       = {Yaowei Zheng and Junting Lu and Shenzhi Wang and Zhangchi Feng and Dongdong Kuang and Yuwen Xiong},
  howpublished = {\url{https://github.com/hiyouga/EasyR1}},
  year         = {2025}
}

@article{Hu2021LoRALA,
  title={LoRA: Low-Rank Adaptation of Large Language Models},
  author={J. Edward Hu and Yelong Shen and Phillip Wallis and Zeyuan Allen-Zhu and Yuanzhi Li and Shean Wang and Weizhu Chen},
  journal={ArXiv},
  year={2021},
  volume={abs/2106.09685},
  url={https://api.semanticscholar.org/CorpusID:235458009}
}

@article{Dettmers2023QLoRAEF,
  title={QLoRA: Efficient Finetuning of Quantized LLMs},
  author={Tim Dettmers and Artidoro Pagnoni and Ari Holtzman and Luke Zettlemoyer},
  journal={ArXiv},
  year={2023},
  volume={abs/2305.14314},
  url={https://api.semanticscholar.org/CorpusID:258841328}
}

@article{Hu2023LLMAdaptersAA,
  title={LLM-Adapters: An Adapter Family for Parameter-Efficient Fine-Tuning of Large Language Models},
  author={Zhiqiang Hu and Yihuai Lan and Lei Wang and Wanyu Xu and Ee-Peng Lim and Roy Ka-Wei Lee and Lidong Bing and Soujanya Poria},
  journal={ArXiv},
  year={2023},
  volume={abs/2304.01933},
  url={https://api.semanticscholar.org/CorpusID:257921386}
}

@inproceedings{Lester2021ThePO,
  title={The Power of Scale for Parameter-Efficient Prompt Tuning},
  author={Brian Lester and Rami Al-Rfou and Noah Constant},
  booktitle={Conference on Empirical Methods in Natural Language Processing},
  year={2021},
  url={https://api.semanticscholar.org/CorpusID:233296808}
}

@article{Li2021PrefixTuningOC,
  title={Prefix-Tuning: Optimizing Continuous Prompts for Generation},
  author={Xiang Lisa Li and Percy Liang},
  journal={Proceedings of the 59th Annual Meeting of the Association for Computational Linguistics and the 11th International Joint Conference on Natural Language Processing (Volume 1: Long Papers)},
  year={2021},
  pages={4582-4597},
  url={https://api.semanticscholar.org/CorpusID:230433941}
}

@article{Shao2024DeepSeekMathPT,
  title={DeepSeekMath: Pushing the Limits of Mathematical Reasoning in Open Language Models},
  author={Zhihong Shao and Peiyi Wang and Qihao Zhu and Runxin Xu and Jun-Mei Song and Mingchuan Zhang and Y. K. Li and Yu Wu and Daya Guo},
  journal={ArXiv},
  year={2024},
  volume={abs/2402.03300},
  url={https://api.semanticscholar.org/CorpusID:267412607}
}

@article{Yu2025DAPOAO,
  title={DAPO: An Open-Source LLM Reinforcement Learning System at Scale},
  author={Qiying Yu and Zheng Zhang and Ruofei Zhu and Yufeng Yuan and Xiaochen Zuo and Yu Yue and Tiantian Fan and Gaohong Liu and Lingjun Liu and Xin Liu and Haibin Lin and Zhiqi Lin and Bole Ma and Guangming Sheng and Yuxuan Tong and Chi Zhang and Mofan Zhang and Wang Zhang and Hang Zhu and Jinhua Zhu and Jiaze Chen and Jiangjie Chen and Chengyi Wang and Honglin Yu and Weinan Dai and Yuxuan Song and Xiang Wei and Haodong Zhou and Jingjing Liu and Wei Ma and Ya-Qin Zhang and Lin Yan and Mu Qiao and Yong-Xu Wu and Mingxuan Wang},
  journal={ArXiv},
  year={2025},
  volume={abs/2503.14476},
  url={https://api.semanticscholar.org/CorpusID:277104124}
}

@article{Zheng2025GroupSP,
  title={Group Sequence Policy Optimization},
  author={Chujie Zheng and Shixuan Liu and Mingze Li and Xionghui Chen and Bowen Yu and Chang Gao and Kai Dang and Yuqiong Liu and Rui Men and An Yang and Jingren Zhou and Junyang Lin},
  journal={ArXiv},
  year={2025},
  volume={abs/2507.18071},
  url={https://api.semanticscholar.org/CorpusID:280017753}
}

@inproceedings{Zhang2025OnPolicyRM,
  title={On-Policy RL Meets Off-Policy Experts: Harmonizing Supervised Fine-Tuning and Reinforcement Learning via Dynamic Weighting},
  author={Wenhao Zhang and Yuexiang Xie and Yuchang Sun and Yanxi Chen and Guoyin Wang and Yaliang Li and Bolin Ding and Jingren Zhou},
  year={2025},
  url={https://api.semanticscholar.org/CorpusID:280671636}
}

@article{Wang2025ReinforcementLF,
  title={Reinforcement Learning for Out-of-Distribution Reasoning in LLMs: An Empirical Study on Diagnosis-Related Group Coding},
  author={Hanyin Wang and Zhenbang Wu and Gururaj J. Kolar and Hariprasad Reddy Korsapati and Brian Bartlett and Bryan Hull and Jimeng Sun},
  journal={ArXiv},
  year={2025},
  volume={abs/2505.21908},
  url={https://api.semanticscholar.org/CorpusID:278960217}
}

@inproceedings{lin2004rouge,
  title={Rouge: A package for automatic evaluation of summaries},
  author={Lin, Chin-Yew},
  booktitle={Text summarization branches out},
  pages={74--81},
  year={2004}
}

@inproceedings{Shenfeld2025RLsRW,
  title={RL's Razor: Why Online Reinforcement Learning Forgets Less},
  author={Idan Shenfeld and Jyothish Pari and Pulkit Agrawal},
  year={2025},
  url={https://api.semanticscholar.org/CorpusID:281103647}
}

@inproceedings{zheng2024llamafactory,
  title={LlamaFactory: Unified Efficient Fine-Tuning of 100+ Language Models},
  author={Yaowei Zheng and Richong Zhang and Junhao Zhang and Yanhan Ye and Zheyan Luo and Zhangchi Feng and Yongqiang Ma},
  booktitle={Proceedings of the 62nd Annual Meeting of the Association for Computational Linguistics (Volume 3: System Demonstrations)},
  address={Bangkok, Thailand},
  publisher={Association for Computational Linguistics},
  year={2024},
  url={http://arxiv.org/abs/2403.13372}
}

@article{Rafailov2023DirectPO,
  title={Direct Preference Optimization: Your Language Model is Secretly a Reward Model},
  author={Rafael Rafailov and Archit Sharma and Eric Mitchell and Stefano Ermon and Christopher D. Manning and Chelsea Finn},
  journal={ArXiv},
  year={2023},
  volume={abs/2305.18290},
  url={https://api.semanticscholar.org/CorpusID:258959321}
}

@article{Lambert2025ReinforcementLF,
  title={Reinforcement Learning from Human Feedback},
  author={Nathan Lambert},
  journal={ArXiv},
  year={2025},
  volume={abs/2504.12501},
  url={https://api.semanticscholar.org/CorpusID:277857379}
}

@article{Li2021TowardsAU,
  title={Towards a Unified Foundation Model: Jointly Pre-Training Transformers on Unpaired Images and Text},
  author={Qing Li and Boqing Gong and Yin Cui and D. Kondratyuk and Xianzhi Du and Ming-Hsuan Yang and Matthew Brown},
  journal={ArXiv},
  year={2021},
  volume={abs/2112.07074},
  url={https://api.semanticscholar.org/CorpusID:245131381}
}

@article{Chung2022ScalingIL,
  title={Scaling Instruction-Finetuned Language Models},
  author={Hyung Won Chung and Le Hou and S. Longpre and Barret Zoph and Yi Tay and William Fedus and Eric Li and Xuezhi Wang and Mostafa Dehghani and Siddhartha Brahma and Albert Webson and Shixiang Shane Gu and Zhuyun Dai and Mirac Suzgun and Xinyun Chen and Aakanksha Chowdhery and Dasha Valter and Sharan Narang and Gaurav Mishra and Adams Wei Yu and Vincent Zhao and Yanping Huang and Andrew M. Dai and Hongkun Yu and Slav Petrov and Ed H. Chi and Jeff Dean and Jacob Devlin and Adam Roberts and Denny Zhou and Quoc V. Le and Jason Wei},
  journal={ArXiv},
  year={2022},
  volume={abs/2210.11416},
  url={https://api.semanticscholar.org/CorpusID:253018554}
}

@article{Schulman2017ProximalPO,
  title={Proximal Policy Optimization Algorithms},
  author={John Schulman and Filip Wolski and Prafulla Dhariwal and Alec Radford and Oleg Klimov},
  journal={ArXiv},
  year={2017},
  volume={abs/1707.06347},
  url={https://api.semanticscholar.org/CorpusID:28695052}
}

@inproceedings{Li2021AlignBF,
  title={Align before Fuse: Vision and Language Representation Learning with Momentum Distillation},
  author={Junnan Li and Ramprasaath R. Selvaraju and Akhilesh Deepak Gotmare and Shafiq R. Joty and Caiming Xiong and Steven C. H. Hoi},
  booktitle={Neural Information Processing Systems},
  year={2021},
  url={https://api.semanticscholar.org/CorpusID:236034189}
}

@inproceedings{Singh2022FLAVAUO,
  title={FLAVA: A Foundational Language And Vision Alignment Model},
  author={Rishabh Singh and Amanpreet Singh and Vivek Nair and Yu-Xiong Wang and Laurens van der Maaten and Armand Joulin and Ishan Misra},
  booktitle={ICML},
  year={2022}
}

@inproceedings{Liu2023VisualIT,
  title={Visual Instruction Tuning},
  author={Haotian Liu and Chunyuan Li and Qingyang Wu and Yong Jae Lee},
  booktitle={NeurIPS},
  year={2023}
}

@inproceedings{reimers-2019-sentence-bert,
    title = "Sentence-BERT: Sentence Embeddings using Siamese BERT-Networks",
    author = "Reimers, Nils and Gurevych, Iryna",
    booktitle = "Proceedings of the 2019 Conference on Empirical Methods in Natural Language Processing",
    month = "11",
    year = "2019",
    publisher = "Association for Computational Linguistics",
    url = "http://arxiv.org/abs/1908.10084",
}

@article{DBLP:journals/corr/abs-1911-02116,
  author    = {Alexis Conneau and
               Kartikay Khandelwal and
               Naman Goyal and
               Vishrav Chaudhary and
               Guillaume Wenzek and
               Francisco Guzm{\'{a}}n and
               Edouard Grave and
               Myle Ott and
               Luke Zettlemoyer and
               Veselin Stoyanov},
  title     = {Unsupervised Cross-lingual Representation Learning at Scale},
  journal   = {CoRR},
  volume    = {abs/1911.02116},
  year      = {2019},
  url       = {http://arxiv.org/abs/1911.02116},
  eprinttype = {arXiv},
  eprint    = {1911.02116},
  bibsource = {dblp computer science bibliography, https://dblp.org}
}

@article{Honnibal_spaCy_Industrial-strength_Natural_2020,
author = {Honnibal, Matthew and Montani, Ines and Van Landeghem, Sofie and Boyd, Adriane},
doi = {10.5281/zenodo.1212303},
title = {{spaCy: Industrial-strength Natural Language Processing in Python}},
year = {2020}
}

@article{chen2015microsoft,
  title={Microsoft coco captions: Data collection and evaluation server},
  author={Chen, Xinlei and Fang, Hao and Lin, Tsung-Yi and Vedantam, Ramakrishna and Gupta, Saurabh and Doll{\'a}r, Piotr and Zitnick, C Lawrence},
  journal={arXiv preprint arXiv:1504.00325},
  year={2015}
}

@inproceedings{papineni-etal-2002-bleu,
    title = "{B}leu: a Method for Automatic Evaluation of Machine Translation",
    author = "Papineni, Kishore  and
      Roukos, Salim  and
      Ward, Todd  and
      Zhu, Wei-Jing",
    editor = "Isabelle, Pierre  and
      Charniak, Eugene  and
      Lin, Dekang",
    booktitle = "Proceedings of the 40th Annual Meeting of the Association for Computational Linguistics",
    month = jul,
    year = "2002",
    address = "Philadelphia, Pennsylvania, USA",
    publisher = "Association for Computational Linguistics",
    url = "https://aclanthology.org/P02-1040/",
    doi = "10.3115/1073083.1073135",
    pages = "311--318"
}

@article{Vedantam2014CIDErCI,
  title={CIDEr: Consensus-based image description evaluation},
  author={Ramakrishna Vedantam and C. Lawrence Zitnick and Devi Parikh},
  journal={2015 IEEE Conference on Computer Vision and Pattern Recognition (CVPR)},
  year={2014},
  pages={4566-4575},
  url={https://api.semanticscholar.org/CorpusID:9026666}
}

@inproceedings{anderson2016spice,
  title={Spice: Semantic propositional image caption evaluation},
  author={Anderson, Peter and Fernando, Basura and Johnson, Mark and Gould, Stephen},
  booktitle={European conference on computer vision},
  pages={382--398},
  year={2016},
  organization={Springer}
}

@inproceedings{
gao2025gllava,
title={G-{LL}a{VA}: Solving Geometric Problem with Multi-Modal Large Language Model},
author={Jiahui Gao and Renjie Pi and Jipeng Zhang and Jiacheng Ye and Wanjun Zhong and Yufei Wang and Lanqing HONG and Jianhua Han and Hang Xu and Zhenguo Li and Lingpeng Kong},
booktitle={The Thirteenth International Conference on Learning Representations},
year={2025},
url={https://openreview.net/forum?id=px1674Wp3C}
}

@article{demner2012design,
  title={Design and development of a multimodal biomedical information retrieval system},
  author={Demner-Fushman, Dina and Antani, Sameer and Simpson, Matthew and Thoma, George R},
  journal={Journal of Computing Science and Engineering},
  volume={6},
  number={2},
  pages={168--177},
  year={2012},
  publisher={Demner-Fushman Dina; Antani Sameer; Simpson Matthew; Thoma George R.}
}

@misc{kumar2025llmposttrainingdeepdive,
      title={LLM Post-Training: A Deep Dive into Reasoning Large Language Models}, 
      author={Komal Kumar and Tajamul Ashraf and Omkar Thawakar and Rao Muhammad Anwer and Hisham Cholakkal and Mubarak Shah and Ming-Hsuan Yang and Phillip H. S. Torr and Fahad Shahbaz Khan and Salman Khan},
      year={2025},
      eprint={2502.21321},
      archivePrefix={arXiv},
      primaryClass={cs.CL},
      url={https://arxiv.org/abs/2502.21321}, 
}

@article{chu2025sft,
  title={Sft memorizes, rl generalizes: A comparative study of foundation model post-training},
  author={Chu, Tianzhe and Zhai, Yuexiang and Yang, Jihan and Tong, Shengbang and Xie, Saining and Schuurmans, Dale and Le, Quoc V and Levine, Sergey and Ma, Yi},
  journal={arXiv preprint arXiv:2501.17161},
  year={2025}
}

@article{lai2025med,
  title={Med-r1: Reinforcement learning for generalizable medical reasoning in vision-language models},
  author={Lai, Yuxiang and Zhong, Jike and Li, Ming and Zhao, Shitian and Yang, Xiaofeng},
  journal={arXiv preprint arXiv:2503.13939},
  year={2025}
}

@article{li2025drive,
  title={Drive-R1: Bridging Reasoning and Planning in VLMs for Autonomous Driving with Reinforcement Learning},
  author={Li, Yue and Tian, Meng and Zhu, Dechang and Zhu, Jiangtong and Lin, Zhenyu and Xiong, Zhiwei and Zhao, Xinhai},
  journal={arXiv preprint arXiv:2506.18234},
  year={2025}
}

@inproceedings{duan2024cityllava,
  title={Cityllava: Efficient fine-tuning for vlms in city scenario},
  author={Duan, Zhizhao and Cheng, Hao and Xu, Duo and Wu, Xi and Zhang, Xiangxie and Ye, Xi and Xie, Zhen},
  booktitle={Proceedings of the IEEE/CVF Conference on Computer Vision and Pattern Recognition},
  pages={7180--7189},
  year={2024}
}

@article{dong2025scalable,
  title={Scalable vision language model training via high quality data curation},
  author={Dong, Hongyuan and Kang, Zijian and Yin, Weijie and Liang, Xiao and Feng, Chao and Ran, Jiao},
  journal={arXiv preprint arXiv:2501.05952},
  year={2025}
}

@inproceedings{chen2024efficiency,
  title={Efficiency in Focus: LayerNorm as a Catalyst for Fine-tuning Medical Visual Language Models},
  author={Chen, Jiawei and Yang, Dingkang and Jiang, Yue and Li, Mingcheng and Wei, Jinjie and Hou, Xiaolu and Zhang, Lihua},
  booktitle={Proceedings of the 32nd ACM International Conference on Multimedia},
  pages={3122--3130},
  year={2024}
}

@article{Li2016LearningWF,
  title={Learning without Forgetting},
  author={Zhizhong Li and Derek Hoiem},
  journal={IEEE Transactions on Pattern Analysis and Machine Intelligence},
  year={2016},
  volume={40},
  pages={2935-2947},
  url={https://api.semanticscholar.org/CorpusID:4853851}
}

@article{Kirkpatrick2016OvercomingCF,
  title={Overcoming catastrophic forgetting in neural networks},
  author={James Kirkpatrick and Razvan Pascanu and Neil C. Rabinowitz and Joel Veness and Guillaume Desjardins and Andrei A. Rusu and Kieran Milan and John Quan and Tiago Ramalho and Agnieszka Grabska-Barwinska and Demis Hassabis and Claudia Clopath and Dharshan Kumaran and Raia Hadsell},
  journal={Proceedings of the National Academy of Sciences},
  year={2016},
  volume={114},
  pages={3521 - 3526},
  url={https://api.semanticscholar.org/CorpusID:4704285}
}

@inproceedings{NEURIPS2023_cdd06402,
 author = {MA, XIAOSONG and ZHANG, Jie and Guo, Song and Xu, Wenchao},
 booktitle = {Advances in Neural Information Processing Systems},
 editor = {A. Oh and T. Naumann and A. Globerson and K. Saenko and M. Hardt and S. Levine},
 pages = {65252--65264},
 publisher = {Curran Associates, Inc.},
 title = {SwapPrompt: Test-Time Prompt Adaptation for Vision-Language Models},
 url = {https://proceedings.neurips.cc/paper_files/paper/2023/file/cdd0640218a27e9e2c0e52e324e25db0-Paper-Conference.pdf},
 volume = {36},
 year = {2023}
}

@article{Niu2023TowardsST,
  title={Towards Stable Test-Time Adaptation in Dynamic Wild World},
  author={Shuaicheng Niu and Jiaxiang Wu and Yifan Zhang and Zhiquan Wen and Yaofo Chen and Peilin Zhao and Mingkui Tan},
  journal={ArXiv},
  year={2023},
  volume={abs/2302.12400},
  url={https://api.semanticscholar.org/CorpusID:257206115}
}

@article{Yang2024AVF,
  title={A Versatile Framework for Continual Test-Time Domain Adaptation: Balancing Discriminability and Generalizability},
  author={Xu Yang and Xuan Chen and Moqi Li and Kun-Juan Wei and Cheng Deng},
  journal={2024 IEEE/CVF Conference on Computer Vision and Pattern Recognition (CVPR)},
  year={2024},
  pages={23731-23740},
  url={https://api.semanticscholar.org/CorpusID:271691678}
}

@article{Rebuffi2016iCaRLIC,
  title={iCaRL: Incremental Classifier and Representation Learning},
  author={Sylvestre-Alvise Rebuffi and Alexander Kolesnikov and G. Sperl and Christoph H. Lampert},
  journal={2017 IEEE Conference on Computer Vision and Pattern Recognition (CVPR)},
  year={2016},
  pages={5533-5542},
  url={https://api.semanticscholar.org/CorpusID:206596260}
}
